\documentclass[letterpaper, 10pt, conference]{ieeeconf} 
\IEEEoverridecommandlockouts 
\usepackage{cite}
\usepackage{amsmath,amssymb,amsfonts}
\usepackage{algorithmic}
\usepackage{graphicx}
\usepackage{textcomp}
\usepackage{xcolor}
\usepackage{booktabs}
\usepackage{siunitx}
\usepackage{comment}
\def\BibTeX{{\rm B\kern-.05em{\sc i\kern-.025em b}\kern-.08em
    T\kern-.1667em\lower.7ex\hbox{E}\kern-.125emX}}

\title{Controllable Diverse Sampling for Diffusion Based Motion Behavior Forecasting}

\author{
Yiming Xu$^{1}$,
Hao Cheng$^{2}$,
Monika Sester$^{1}$
\thanks{$^{1}$,Yiming Xu and Monika Sester are with the Institute of Cartography and Geoinformatics, Leibniz University Hannover, Appelstr. 9a, 30167 Hannover, Germany {\tt\small\{Yiming.Xu, Monika.Sester\}@ikg.uni-hannover.de}}%
\thanks{$^{2}$,Hao Cheng is with the Faculty of Geo-Information Science and Earth Observation, University of Twente, 7500 AE Enschede, The Netherlands. {\tt\small h.cheng-2@utwente.nl}}
}

\begin{document}
\maketitle

\begin{abstract}
In autonomous driving tasks, trajectory prediction in complex traffic environments requires adherence to real-world context conditions and behavior multimodalities. Existing methods predominantly rely on prior assumptions or generative models trained on curated data to learn road agents' stochastic behavior bounded by scene constraints. However, they often face mode averaging issues due to data imbalance and simplistic priors, and could even suffer from mode collapse due to unstable training and single ground truth supervision. These issues lead the existing methods to a loss of predictive diversity and adherence to the scene constraints. To address these challenges, we introduce a novel trajectory generator named Controllable Diffusion Trajectory (CDT), which integrates map information and social interactions into a Transformer-based conditional denoising diffusion model to guide the prediction of future trajectories. To ensure multimodality, we incorporate behavioral tokens to direct the trajectory's modes, such as going straight, turning right or left. Moreover, we incorporate the predicted endpoints as an alternative behavioral token into the CDT model to facilitate the prediction of accurate trajectories. Extensive experiments on the Argoverse~2 benchmark demonstrate that CDT excels in generating diverse and scene-compliant trajectories in complex urban settings.
% , and achieves remarkable accuracy in trajectory generation.
\end{abstract}

\section{Introduction}
% At present, addressing trajectory prediction problems primarily involves data-driven approaches that interact sequential trajectories with map scene context in a latent space, subsequently learning the potential multimodal possibilities and trajectory representations for the future. In this process, the multimodal output is particularly crucial for trajectory prediction.

% For autonomous vehicle systems, accurately foreseeing the potential action trajectories of nearby vehicles is vital for interpreting the surrounding environment and making effective decisions. 
The goal of realizing multimodal trajectory prediction is to generate a series of potential trajectories for the observed vehicles, which plays an essential role in addressing uncertainties in driving behavior prediction and enhancing the safety of path planning. 
% However, due to the unpredictability of the future, different vehicles may exhibit various behavior patterns under the same driving scenario. 
% However, typically, only one actual driving trajectory is available in each driving environment.
To train such a trajectory prediction model that can effectively predict the multifaceted future behaviors of the observed vehicles requires a substantial amount of diverse real trajectories as training samples in similar scenarios. 
However, in existing datasets, data samples are often biased.
For example, in Argoverse~1 \cite{chang2019argoverse}, the data proportions for straight, left turn, and right turn are 83.6\%, 10.2\%, and 6.2\%, respectively; Similarly, in Argoverse~2 \cite{wilson2023argoverse}, these proportions are 81.7\%, 9.5\%, and 8.8\%.
It can be seen that the number of straight trajectories far exceeds that of left or right turns.
This likely leads a model to learn a higher probability of vehicles moving straight, thereby generating less accurate and diverse future trajectories.
% Therefore, one of the primary challenges in achieving diverse trajectory prediction is covering all possible driving schemes in a scenario with limited training samples.
% Moreover, collecting a vast amount of real-world trajectory data is a significant challenge.  However, in existing datasets, the imbalance in data of different modal trajectories may impact the prediction accuracy and robustness of autonomous driving systems.
% In existing datasets, such as Argoverse1 \cite{chang2019argoverse} and Argoverse~2 \cite{wilson2023argoverse}, the number of straight trajectories far exceeds that of left or right turns. This leads data-driven methods to learn a higher probability of vehicles moving straight, thereby increasing the inaccuracy of trajectory prediction probabilities.  Based on these datasets, 

To battle imbalanced data samples, many existing models primarily rely on probabilistic methods based on specific prior distributions. 
For example, Gaussian Mixture Models \cite{reynoids2009} are widely explored to sample diverse futures for trajectory prediction.
However, these methods heavily rely on carefully designed loss functions and training strategies (e.g., winner-takes-all)  to mitigate the mode collapse problem \cite{makansi2019overcoming, zhou2022hivt, deo2022multimodal, liu2023laformer}. The same results are output under different scene information and maps.
Moreover, generative models such as Generative Adversarial Networks (GANs) \cite{gupta2018social, sadeghian2019sophie}, Conditional Variational Autoencoders (CVAEs) \cite{mangalam2020not, xu2022socialvae}, and Flow models \cite{liang2023stglow} are employed for multimodal trajectory prediction. 
However they are known to suffer from the risk of mode averaging problems \cite{Richardson2018} when trained on a single ground-truth trajectory. Only the average or typical representation is output and all possibilities are averaged.
% Direct modeling of all trajectory data based on simple distribution assumptions (i.e., Gaussian distribution) somewhat limits the model's expressive capability, leading to mode average problems in autonomous driving scenarios \cite{zhao2021tnt}. Moreover, the predominance of straight-driving data in datasets results in a disproportionately high probability of sampling straight trajectories at intersections compared to other possibilities. Both mode collapse and mode average problems lead to reduced predicted trajectory diversity.

\begin{figure}[t!]
\centerline{\includegraphics[width=8.5cm]{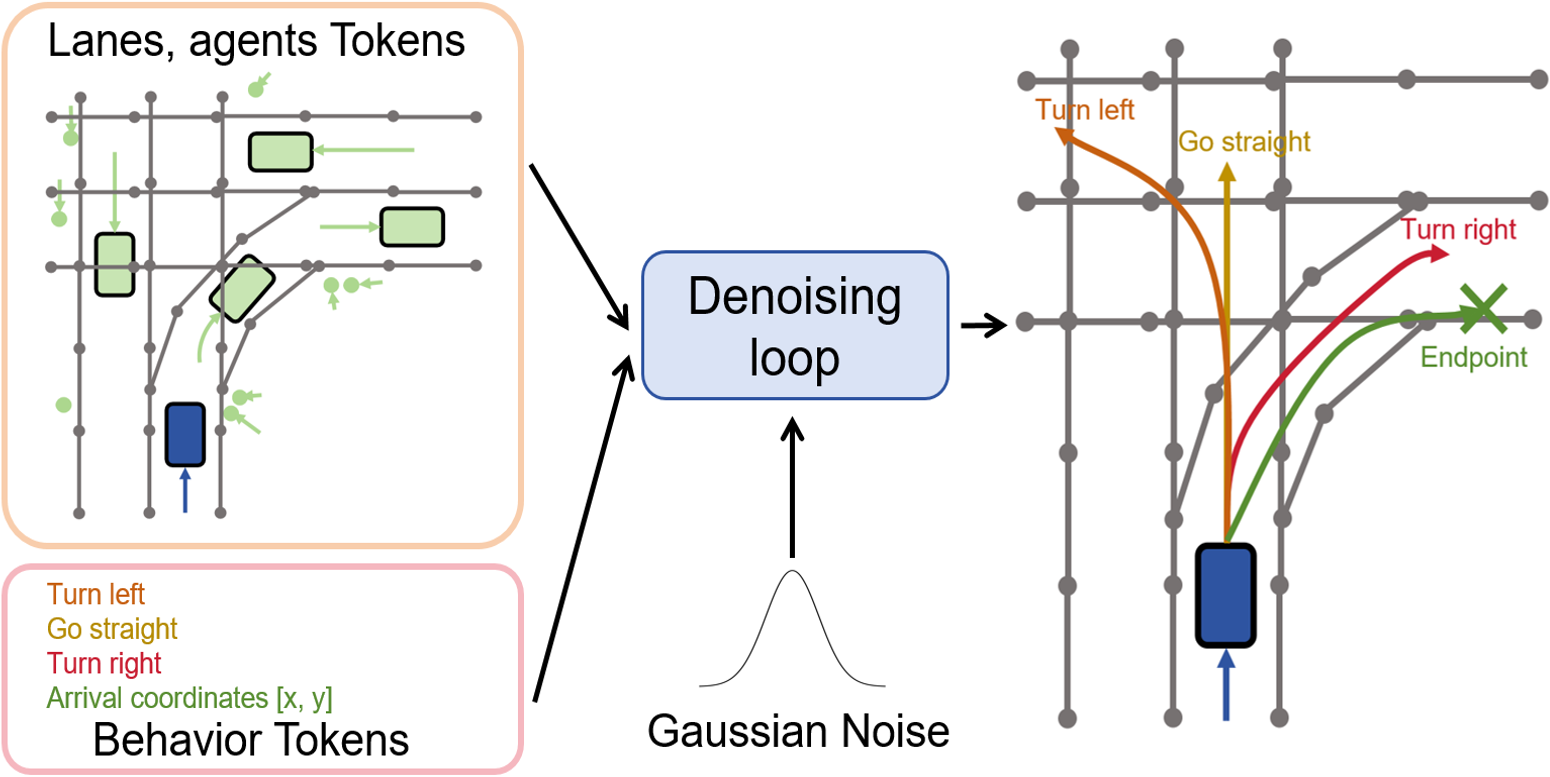}}
\caption{CDT leverages traffic condition information and behavioral tokens as denoising conditions, sampling condition-based trajectories from Gaussian noise.}
\vspace{-6pt}
\label{SchematicDiagram}
\end{figure}

In this work, we propose a novel multimodal trajectory prediction framework, called Controllable Diffusion Trajectory (CDT) illustrated in Fig. \ref{SchematicDiagram}, that controllably generates trajectories of different modalities. 
The CDT model leverages the excellent feature extraction capability of conditional denoising diffusion model \cite{ho2020denoising, choi2021ilvr} to generate controllable multimodal trajectories.
Concretely, CDT introduces a mode classifier that estimates the target vehicle's behavior based on the road information, guiding the diffusion model to sample the most probable trajectories compliant with the scene constraints.
Moreover, CDT utilizes a transformer-based Denoiser to include motion behavior control, i.e., \textit{straight}, \textit{left turn}, and \textit{right turn}, or \textit{endpoints}, as behavior tokens with the agent's historical trajectory token, lane token, and time embedding.
Consequently, the Denoiser denoises the final predictions conditioned on the combined tokens.  
This way, CDT can easily avoid bias towards the majority behavior patterns in predictions and synthesize diverse and scene-compliant trajectories.
Compared to other existing models, CDT can provide more diverse, high-quality trajectories even under data imbalance conditions.
% and one-dimensional convolutions for sequence modeling, utilizing their strong expressive power for complex problems and multimodality, as well as their condition-based controllability to generate desired motion behavior trajectories. We employ the denoising diffusion model to model all trajectory suggestions, learning the diverse possibilities of future scenarios under different driving conditions. Our diffusion model includes motion behavior control based on Transformer-decoder \cite{vaswani2017attention}, namely straight, left turn, and right turn or end position. When facing data imbalance issues, our condition-based modeling approach effectively avoids bias towards the majority behavior pattern (straight driving) in predictions. We also introduce a behavior classification module, which estimates the target vehicle's behavior based on road information, guiding the diffusion model to sample the most probable paths. Simultaneously, we can synthesize a large number of different behavioral modal trajectories in the same scenario that comply with traffic behavior rules, which is crucial for analyzing the behavior of target vehicles at traffic intersections. 

In summary, our contributions are as follows:

\begin{itemize}
\item We propose a novel multimodal motion prediction sampling framework that employs a conditional denoising diffusion model with a Transformer-based decoder. 
% It effectively addresses complex trajectory prediction issues and generates various possible behavior patterns. 
% This method excels in handling data imbalance issues, avoiding excessive bias towards dominant behavior patterns in the data, thereby enhancing prediction diversity and simply and effectively avoiding mode collapse and mode average problems.
\item The proposed model is able to synthesize a large number of controllable diverse trajectories by taking specific behavior tokens. 
% data under data imbalance conditions in the same intersection scenario. 
These synthesized data help to increase data sample diversity, especially for less common behavior patterns.
\item The proposed model demonstrates superior performance in the Argoverse~2 dataset.
It achieves excellent results in generating multimodal trajectories with high diversity and scene compliance.
\end{itemize}

\section{Related Works}
\subsection{Motion and Scene Encoding}
The challenges of motion prediction primarily involve the need for an agent (e.g., vehicle or pedestrian) to learn representations and interactions of other agents within the complex traffic environments to avoid collisions or conflicts \cite{pellegrini2009you,klugl2007large,liu2023b}.
To this end, high-definition maps are commonly utilized to represent the scene constraints, and the past trajectories of all dynamic targets are leveraged for agent-to-agent interaction modeling. 
Current leading methods employ vectorized scenes for map encoding, achieving remarkable results \cite{zhao2021tnt, zhou2022hivt, liu2023laformer, gilles2021thomas, feng2023macformer, wang2023prophnet, gao2023dynamic, zhou2023query, varadarajan2022multipath++}.
Compared to rasterized scene representations, such as images or semantic maps \cite{bansal2018chauffeurnet, chai2019multipath, cui2019multimodal, salzmann2020trajectron++, phan2020covernet, gilles2021home}, vectorized high-definition map information can be encoded in the same sequential format as trajectories and aligned with the agents' dynamic information.
For example, VectorNet \cite{gao2020vectornet} proposes a unified vectorized form represented by points, polylines, and polygons for trajectories and scene context. This not only reduces the computational resources required by the encoder, enhancing the model's map perception capabilities, but also simplifies the fusion of spatial information from high-definition maps and temporal information from agent trajectories through the same encoder.
In this work, we adopt the same encoding scheme with vectorized map information for motion and scene information encoding.
% Numerous deep learning-based works employ rasterized scenes as model inputs \cite{bansal2018chauffeurnet, chai2019multipath, cui2019multimodal, salzmann2020trajectron++, phan2020covernet, gilles2021home}, with high-definition maps and past trajectories fed into the model as multi-channel images, processed through image models based on convolutional neural networks. However, rasterized maps, due to their need for extensive perceptual filters and computational resources in handling scene context, and limited receptive fields, may not accurately analyze road features in complex intersections, leading to discrepancies between predicted and actual trajectories. To address this, VectorNet \cite{gao2020vectornet} proposes a unified expression of high-definition maps and agent trajectories in a vectorized form, represented by points, polylines, and polygons for trajectories and scene context. This not only reduces the computational resources required by the encoder, enhancing the model's map perception capabilities, but also simplifies the fusion of spatial information from high-definition maps and temporal information from agent trajectories through the same encoder.  We will also use more information-dense vectorized scenes as inputs to the model, with each segment of the drivable area being used as guidance for the diffusion model to generate trajectories that are more consistent with the actual map.

\subsection{Multimodal Trajectory Decoding}
To address the uncertainties in agent behavior in complex environments and to consider various possibilities, a multimodal trajectory decoder is widely applied for trajectory prediction.
For example, generative models such as Generative Adversarial Networks (GANs) \cite{gupta2018social, sadeghian2019sophie}, Conditional Variational Autoencoders (CVAEs) \cite{mangalam2020not, xu2022socialvae}, and Flow models \cite{liang2023stglow} are commonly employed.
However they face the risk of modal average problems when trained on a single ground-truth trajectory.
\cite{makansi2019overcoming, zhou2022hivt, deo2022multimodal, liu2023laformer} employ multiple prediction heads and train different modalities using a winner-takes-all strategy.
However, similarly, aligning multimodal predictions with the ground truth trajectory of a single modality presents a significant challenge and often leads to unstable training or even mode collapse.
Moreover, due to the imbalance in training data with dominant straightforward motion patterns, the likelihood of predicting straightforward trajectories is higher than that of other modalities.
To overcome these issues, mmTransformer \cite{liu2021multimodal} applies masked methods to constrain different modalities, effectively preventing mode collapse and mode average problems. 
Inspired by this method, we explore the diffusion models with their excellent multimodal modeling capabilities conditioned on specific behavior tokens to further ensure controllability in the multimodal sampling process.
By explicitly encoding different modalities into the conditions of the diffusion model, we obtain multimodality while generating high-quality trajectories.

\begin{figure*}[htbp!]
\centerline{\includegraphics[width=17.8cm]{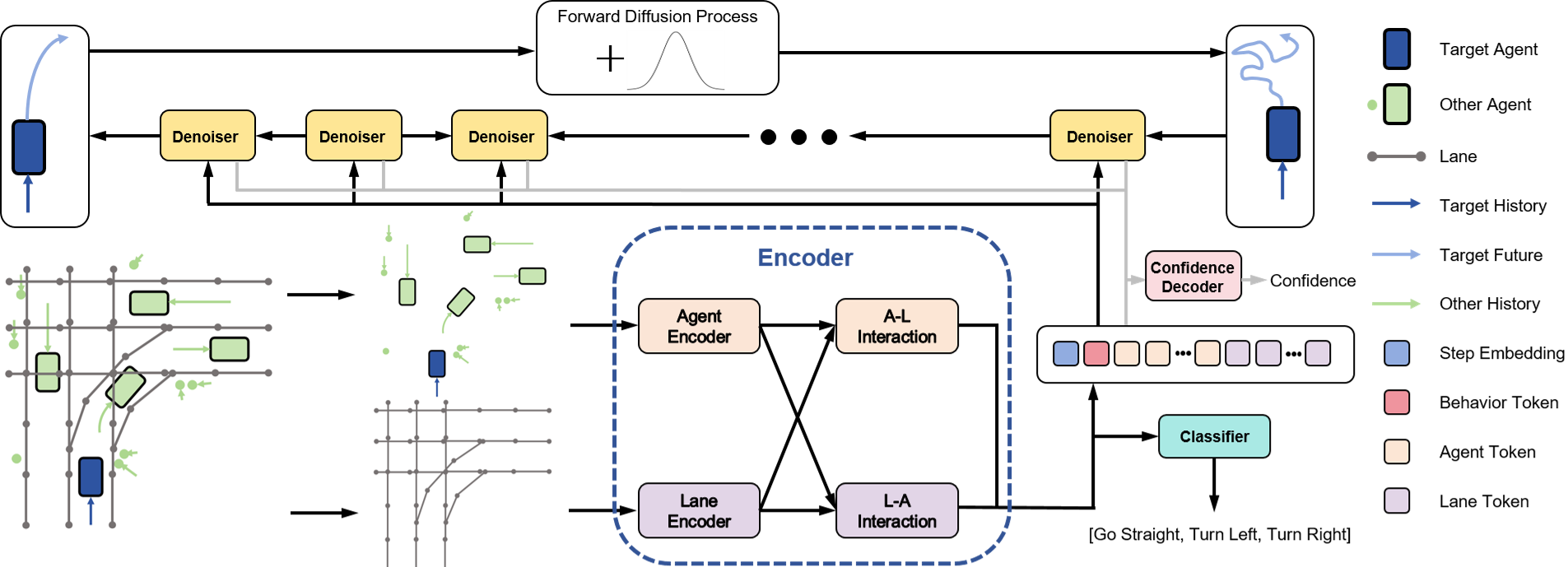}}
\caption{Our trajectory prediction framework utilizes a controllable diffusion model, consisting of an Encoder, a Confidence Decoder, a Classifier, and a transformer-based Denoiser. The Encoder encodes historical trajectories and maps, integrating these with time steps and behavioral conditions as inputs for the denoiser. In the diffusion phase, ground truth trajectories are incrementally degraded by Gaussian noise across T iterations. In the inference phase, The Denoiser iteratively denoises noisy data T steps. The model also outputs future behavior classifications and trajectory confidence levels by Confidence Decoder and Classifier.}
\label{model}
\end{figure*}

\subsection{Conditional Diffusion Models}
Denoising diffusion models, known for their high generation quality and rich diversity, use a Markov-chain forward diffusion process and a step-by-step denoising process to generate desirable data samples \cite{ho2022classifier}.
They have achieved profound success in fields such as image generation \cite{ho2020denoising, ramesh2022hierarchical}, 3D generation \cite{ramesh2022hierarchical, lin2023magic3d}, and video production \cite{ho2022imagen, yang2023diffusion}.
Moreover, diffusion models are used for predicting and simulating human body pose sequences \cite{saadatnejad2023generic, zhang2022motiondiffuse, barquero2023belfusion}.
The application of diffusion models in trajectory prediction is also being explored. 
For example, M. Janner et al. \cite{janner2022planning} utilize diffusion models for planning robot states and joint trajectories, achieving variable-length trajectory planning. 
Jiang et al.~\cite{jiang2023motiondiffuser} employs score-based diffusion models for predicting the joint motion of agents. 
T. Gu et al.~\cite{gu2022stochastic} use diffusion models to forecast future behaviors of pedestrians on streets.
Mao et al.~\cite{mao2023leapfrog} propose a Leapfrog Diffusion to enhance the efficiency of diffusion models in trajectory prediction for real-time forecasting.
However, as generative models, despite their rich diversity, diffusion models can still fall into mode average problems and are influenced by data imbalance issues. 
In recent developments, conditional denoising diffusion models have been extensively researched for their controllability and diversity with classifiers \cite{dhariwal2021diffusion} and without classifiers \cite{ho2022classifier}. 
In this work, we further explore the controllability of diffusion models with a behavior mode classifier, using drivable areas as guidance to ensure trajectory generation aligns with real-world scene constraints, and incorporating simple behavioral conditions (future directions or endpoints) as tokens to guide trajectory generation.

\section{Method}
\subsection{Preliminaries}
Diffusion models are a class of probabilistic models, inspired by non-equilibrium thermodynamics \cite{ho2020denoising}. 
They define a Markov chain forward diffusion process that gradually introduces noise, incrementally degrading the structural integrity of the original data. The forward noise process is defined as follows:
\begin{equation}
\label{forward}
q(\mathcal{T}_t \mid \mathcal{T}_0)=\mathcal{N}(\mathcal{T}_t\mid \sqrt{ \bar{\alpha}_t } \mathcal{T}_0, (1-\bar{\alpha}_t)I),
\end{equation}
which changes the original data $\mathcal{T}_0$ into noisy data $\mathcal{T}_t$ by continuously corrupting the data with Gaussian noise. Here, $t \in \{0, 1, ..., T\}$, $\bar{\alpha}_t=: {\prod_{t}^{t=0}}\alpha_i= {\prod_{t}^{t=0}}(1-\beta_t) $, and $\beta_t$ represents the noise variance schedule \cite{ho2020denoising}. 
The generation process is the inverse of the above process, i.e., the iterative denoising process. 
The diffusion model starts from the standard Gaussian prior $p(\mathcal{T}_T)$ and step-by-step from $\mathcal{T}_T$ to $\mathcal{T}_0$ to recover the original data distribution $p(\mathcal{T}_0)$.
This denoising process is defined as follows:
\begin{equation}
\label{recover}
p(\mathcal{T}_0)=\int p(\mathcal{T}_T )\prod_{T}^{t=1} p_\theta (\mathcal{T}_{t-1} \mid \mathcal{T}_t, C)d\mathcal{T}_{t:T},
\end{equation}
Where $C$ is the condition of the diffusion model to control the results of the mode generation. 

The parameters $\theta$ are optimized by minimizing the variational bound on the negative log-likelihood of the inverse process: $\mathbb{E}_{\mathcal{T}_0} (-log(p_\theta(\mathcal{T}_0)))$. 
Since the computation of $p_\theta (\mathcal{T}_{t-1} \mid \mathcal{T}_i, C)$ is dependent on the entire data distribution, the generation process is usually parameterized as a Gaussian with a fixed time step to predict the mean $\mu_\theta$ and the associated covariance matrix $\Sigma_\theta$:
\begin{equation}
\label{backward}
p_\theta (\mathcal{T}_{t-1} \mid \mathcal{T}_t, C):=\mathcal{N}(\mathcal{T}_{t-1} \mid \mu_\theta(\mathcal{T}_t,t), {\Sigma_\theta}(\mathcal{T}_t,t), C).
\end{equation}

In this work, our objective is to employ conditional diffusion models to address the problem of trajectory generation with controllability. 
Therefore, we define the condition of the proposed diffusion model as the high-definition map information for the guidance of the drivable area and the behavior tokens to control the generated modes. 
Namely, the inputs to this model include information on the past trajectories of other agents (including vehicles, pedestrians, and cyclists), high-definition maps, and behavioral conditions.
% Our primary dataset comprises a future trajectory $\mathcal{T}_0=S$, where $S \in \mathbb{R}^{N \times 2}$, representing a set of two-dimensional coordinate point sequence.
A parameterized network is designed to predict $\mathcal{T}_0$ from random coordinate points $\mathcal{T}_T$, sampled from a Gaussian distribution. 
In our case, $\mathcal{T}_0=S$ is a set of future trajectories characterized by a set of two-dimensional coordinate point sequences, where $S \in \mathbb{R}^{N \times 2}$.
The network also outputs the probabilities of different behavior modes $B$, and a trajectory confidence score $\text{Conf}$. The overview of the framework is shown in Fig.~\ref{model}.

\subsection{Overall Architecture}
The conditional denoising model is bifurcated into two major components: a conditional encoder and a denoising generator.
During the training phase, both parts are trained jointly in an end-to-end manner. 
In the inference phase, the conditional encoder only needs to be executed once as we employ a straightforward Transformer-based encoder (more details in the following subsection).
Whereas, the generator iterates $T$ times for each of the complete denoising chain process. 
To achieve multimodal generation by generating $k$ controllable samples specified by the behavior tokens, the generator operates $k \times T$ times. 
% In this work, the conditional encoder is further divided into two parts: guidance, which includes the past trajectories of all agents and high-definition maps, and behavioral conditions.

\vspace{3pt}
\noindent\textbf{Encoding.}
In our proposed CDT model, we design an attention-based interaction network to encode the past motion and map information of each agent.
Similarly to VectorNet \cite{gao2020vectornet}, for each agent's trajectory information $A_i$, we use Multi-Layer Perceptrons (MLP) and a Gated Recurrent Unit (GRU) layer to encode the temporal series information. 
A self-attention network layer and a feed-forward layer are employed to encode the map information $M_i$ composed of vectorized discrete points. 
To be more specific, in a given scene, each agent and each segment of the drivable lane are encoded into the representations $a_i$ for $i \in \{1, 2, ..., N_\text{agent}\}$ and $m_j$ for $j \in \{1, 2, ..., N_\text{lane}\}$, respectively. 
Similar to\cite{zhou2022hivt}, we utilize cross-attention mechanisms to integrate lane information with agent information, which are represented as A-L and L-A shown in Fig.~\ref{model}. The latent representations of each surrounding agent and drivable map segment are used as the agent token and the map token as conditions for the Denoiser.
%This process involves:
%\begin{equation}
%\label{A2L}
%a_i = a_i + CrssAttn\{a_i, m_j\} for j \in \{1, 2, ..., N_{lane}\}
%\end{equation}
%\begin{equation}
%\label{L2A}
% m_j =  m_j + CrssAttn\{m_j, a_i\} for i \in \{1, 2, ..., N_{agent}\}
%\end{equation}
% All of the above fused agent and scene information will be used as a guide for the conditional diffusion model to control the generation of future trajectories.

% \subsubsection{Multimodal Decoding}
\vspace{3pt}
\noindent\textbf{Mode classifier.}
Unlike most of the prediction models, e.g., \cite{mangalam2020not, xu2022socialvae,zhou2022hivt}, that directly decode the future trajectory of the target agent from the encoded information, we design a mode classifier to predict the behavior modes to facilitate the estimation of prediction confidence.
We categorize driving behaviors into three types: \textit{going straight}, \textit{turning left}, and \textit{turning right}. 
Due to the limited occurrences of other driving behaviors in the dataset, such as U-turns and parking or standing still, they are insufficient for effective model training.
In this work, we treat U-turns as a type of turning.
Since parking or standing still does not involve any turning movements, we 
merge parking or standing still and going straight together. 
It should be noted that the classifier can be easily extended to include these less common behavior modes as separate classes.
In our model, the outputs of the Encoder, comprising information about all agents' motion dynamics and map information, are first processed through a self-attention layer for interactions between agents. This processed information is then combined with the trajectory representation of the ego agent itself. 
Subsequently, MLPs are employed to output the probabilities of various behavior modes that are compliant with the scene constraints.

\vspace{3pt}
\noindent\textbf{Confidence decoder.}
% In our model, the outputs of the Guidance Encoder, comprising information about all agents and the map, are first processed through a self-attention layer. This processed information is then combined with the trajectory representation of the ego agent itself. Subsequently, MLPs are employed to output the probabilities of various behaviors.
We use a confidence decoder to estimate the confidence of each generated trajectory for the multimodal prediction.
The confidence is determined by processing the intermediate layers of the diffusion model through MLPs, which generates a confidence score for the generated trajectory. 
Specifically, the confidence for each trajectory is computed at each denoising step of the iteration by evaluating the trajectory $p(\mathcal{T}_t)$ obtained at that denoising step \cite{ho2020denoising}. This evaluation is achieved through the simple operations of the MLPs that estimate the difference of the probability distribution of the current $p(\mathcal{T}_t)$ and the true distribution $p(\mathcal{T}_0)$, which is used L1loss to control the estimated difference ($\mathcal{L}_{Conf}$).
To account for the predicted behavior mode with the generated trajectory, the final confidence score is the product of the classification score of the mode classifier and the score derived from the confidence decoder.
% To estimate the confidence of the trajectory, we calculate the mean of the difference between the true distribution and the estimated distribution $\mathcal{L}_F$, which is L1Loss. This approach allows us to assess the accuracy of the trajectory generated by the model by comparing it with the actual trajectory data. By quantifying the disparity between these distributions, the model provides a measure of confidence or reliability for each predicted trajectory, indicating how closely the model's output aligns with real-world trajectory patterns.

\vspace{3pt}
\noindent\textbf{Behavior tokens.}
% We categorize driving behaviors into three types: \textit{going straight}, \textit{turning left}, and \textit{turning right}. 
% Due to the limited occurrences of other driving behaviors in the dataset, such as U-turns and parking, they are insufficient for effective model training.
% In this work, we treat U-turns as a type of turning.
% Since parking or standing still does not involve and turning movements, we 
% merge parking or standing still and going straight together. 
% It should be noted that these less common behavior modes can be easily extended in our proposed behavior tokens.
% Based on the mode classifier, we use the prediction of the three driving behaviors as the behavior tokens.
To diversify the sampling process and avoid biased behavior, we add three behavior modes, \textit{going straight}, \textit{turning left}, and \textit{turning right} as behavior tokens to control the generator.
In this way, no matter how the data samples are biased, e.g., going straight, these behavior tokens can still ensure that the generator produces other modalities, especially at intersections.
In this setting, we call our model CDT-B (Behavior).
We follow the benchmark of Argoiverse~2 \cite{wilson2023argoverse} to set the modality number as six for multimodal trajectory prediction.
Namely, in the inference time, we sample six trajectories with the same behavior in non-intersection scenarios as the scene context prefers straightforward driving behavior, whereas we sample two for each of the three behavior modes at cross-intersections, forcing the generator to search for different driving directions equally.
Alternatively, we incorporate a predicted endpoint as an informative prior into the diffusion model's conditional input, in order to generate accurate predictions based on a highly likely endpoint.
In this setting, our model is called CDT-P (Point).
In this work, given its leading performance on the Argoverse 2 leaderboard, we adopt the most recent query-based model QCNet\cite{zhou2023query} for endpoint prediction.
% Our CDT model becomes a goal-based planning modelin this setting, and it is called CDT-P.
We treat our CDT model that does not take any behavior tokens as the baseline model.
In this work, we use MLPs to encode all these behavior tokens to control the generation of future trajectories.

% I need to discuss with you, then I know how to edit here.
\vspace{3pt}
\noindent\textbf{Diffusion for trajectory prediction.}
Finally, we employ a Transformer-based denoising decoder, aimed at effectively learning the relationships between each future time step and every segment of the drivable area on the map, as well as the behavioral conditions encoded in the behavior tokens, so as to generate more realistic future trajectories. 
Concretely, in each layer of our denoising module, we incorporate a one-dimensional convolutional layer to process the temporal information, a cross-attention mechanism to learn the interactions across different tokens, including lane token, agent token, behavior token, as well as the step embedding that contains the ordered step information of the denoising chains.
In the end, we employ another self-attention mechanism to facilitate the decoding of future trajectories. 
In total, the decoder utilizes six such denoising modules.
The code with detailed structure of our model will be published after the acceptance of this paper. 

\subsection{Training}
During the training process, we first establish a diffusion process from ground truth trajectories to noisy trajectories, and then train the model to reverse this process. For the task of trajectory generation, our loss function can be simplified to an $\mathcal{L}2$ loss \cite{ho2020denoising} as the regression loss $\mathcal{L}_\text{reg}$. This is achieved by controlling the mean of the output distribution of the model through the calculation of the $\mathcal{L}2$ distance between the noise outputted by the model $\epsilon '$ and the noise used in constructing the noisy trajectories $\epsilon$:
\begin{equation}
\label{Lreg}
 \mathcal{L}_\text{reg} = \left \| \epsilon ' - \epsilon   \right \| _2
\end{equation}
For the task of behavior classification, we employ cross-entropy loss to control the loss. 
% This loss function is particularly effective for classification problems as it quantifies the difference between two probability distributions - the predicted probability distribution outputted by the model $p_c$ and the actual distribution $y_c$ of the labels. 
The cross-entropy loss is calculated as follows:
\begin{equation}
\label{Lclass}
 \mathcal{L}_\text{class} = -\sum_c y_c \log(p_c) \ \text{for}\  c \in \{\text{straight}, \text{right},\text{left}\},
\end{equation}
where $p_c$ is the predicted probability distribution outputted by the model and $y_c$ is the actual distribution of the labels.

The total loss is given by:
\begin{equation}
\label{Loss}
 \mathcal{L} = \mathcal{L}_\text{reg} + \gamma_1\mathcal{L}_\text{class} + \gamma_2\mathcal{L}_{Conf}
\end{equation}
where $\gamma_1$ and $\gamma_2$ are the hyperparameters used to balance between different loss functions.

\section{Experiments}
\subsection{Experimental Setup}
\subsubsection{Dataset}
We evaluated our prediction framework on the large-scale Argoverse~2 motion prediction dataset \cite{wilson2023argoverse}, which provides trajectories of agents and high-definition map data. Argoverse~2 consists of 25,000 real driving scenarios, divided into training, validation, and test sets with 199,908, 24,988, and 24,984 samples, respectively. Training and validation scenes in Argoverse~2 are 11-second sequences sampled at \SI{10}{Hz}, with only the first five seconds of trajectory data available in the test set. Given the first five seconds of observation, our model predicts the agent's motion for the subsequent six seconds.

\subsubsection{Metrics}
For evaluating the prediction performance, we are more focused on the diversity and scene compliance of the generated trajectories.
Hence, we use Average Self Distance (ASD) \cite{yuan2019diverse} and Expected Collision-Free Likelihood (ECFL) \cite{sohn2022a2x} to measure the performance for diversity and scene compliance, respectively.
\begin{itemize}
    \item ASD$_K$ and FSD$_K$ measure the pairwise Final Displacement Error (FDE) and Average Displacement Error (ADE) among the $K$ generated trajectories. Higher values for these metrics indicate greater diversity in the generated trajectories. Here, $K$ refers to the number of modes. FDE and ADE measure the $\mathcal{L}2$ error at the final step and average at each step between the prediction trajectory and the corresponding ground truth trajectory.
    \item ECFL calculates the probability of generated trajectories avoiding collisions with objects, as proposed in \cite{sohn2022a2x}. In this work, it is tailored to measure the likelihood of avoiding non-drivable areas defined by the high-definition map. A higher ECFL$_K$ close to 1 is more desirable, indicating better scene compliance.
    \item Moreover, we also follow the minFDE$_K$ and minADE$_K$ to measure the $\mathcal{L}2$ error at the final step and average at each step, respectively, for predicting $K$ modes. Here, we report the minimum error across $K$ modes. Both minADE and minFDE are measured in meters. 
    \item Additionally, the Miss Rate (MR$_K$) measures the percentage of scenarios where the error in the final step exceeds 2.0 meters. 
\end{itemize}
Higher values of ASD$_K$ and FSD$_K$ indicate richer diversity. For other evaluation metrics, lower values are preferable.

% For diversity, we use Average Self Distance (ASD$_K$) and Final Self Distance (FSD$_K$) to measure the pairwise FDE and ADE among the $K$ generated trajectories. Higher values for these metrics indicate greater diversity in the generated trajectories. 

% In our study, we quantified the diversity of trajectories on the validation set using Average Self Distance (ASD) and Future Self Distance (FSD). 
% % This evaluation metric has to be further explained.

% Furthermore, in our ablation studies to measure the congruence of generated trajectories with the map, we employ the Expected Collision-Free Likelihood (ECFL$_K$) \cite{sohn2022a2x} metric. This metric calculates the probability of generated trajectories avoiding collisions with objects; in this work, it is tailored to measure the likelihood of avoiding non-drivable areas. 
% A higher ECFL$_K$ is desirable.

% In Argoverse~2, we set $K$ to 6. These above metrics collectively provide a comprehensive assessment of the model's accuracy, reliability, and diversity in trajectory prediction.

\subsubsection{Implementation Details}
We trained our model using the AdamW optimizer \cite{loshchilov2017decoupled} on an RTX 3060 GPU. The settings for epochs, batch size, and initial learning rate were configured to 140, 64, and $5\times 10^{-4}$, respectively. Our training strategy for the learning rate involved an initial warmup phase of 1500 steps, followed by cosine annealing scheduler \cite{loshchilov2016sgdr} for learning rate decay. The number of heads in all multi-head attention blocks was set to 4. We did not employ techniques such as ensemble methods or data augmentation. Our conditional diffusion model utilized 20 denoising steps.
\subsection{Results}

\begin{comment}
\begin{table}[!t]
\caption{In the test set, quantitative results of generated trajectories and compared with state-of-the-art methods.}
\begin{center}
\begin{tabular}{l|ccc}
\toprule
\textbf{Method}&\textbf{minFDE$_6$}&\textbf{minADE$_6$}&\textbf{minMR$_6$} \\
\midrule
THOMAS\cite{gilles2021thomas} & 1.51 & 0.88 & 0.20 \\
MacFormer\cite{feng2023macformer} & 1.38 & 0.70 & 0.19  \\ 
ProphNet\cite{wang2023prophnet} & 1.32 & 0.66 & 0.18  \\ 
Gnet\cite{gao2023dynamic}  & 1.34 & 0.69 & 0.18  \\
QCNet\cite{zhou2023query} & 1.19 & 0.62 & 0.14  \\
\midrule
CDT-B (Ours) & 2.97 &  1.33 & 0.47  \\
CDT-P (Ours) & 1.25 &  0.80 & 0.16  \\
\bottomrule
\end{tabular}
\label{tab1}
\end{center}
\end{table}
\begin{table}[!t]
\caption{Comparison of trajectory generation diversity in the validation set.}
\begin{center}

\begin{tabular}{l|cc}
\toprule
\textbf{Method}&\textbf{ASD$_6$}&\textbf{FSD$_6$} \\
\midrule
QCNet\cite{zhou2023query} & 8.45 & 8.46 \\
CDT-B (Ours) & 14.49 &  40.75 \\
CDT-NoBehavior (Ours) & 11.33 & 18.02 \\
\bottomrule
\end{tabular}
\label{tab2}
\end{center}
\end{table}
\end{comment}

\vspace{3pt}
\noindent\textbf{Quantitative results.}
Table \ref{tab3} presents the performances of our proposed model with different settings.
First, compared to the baseline model CDT, CDT-B with three driving behaviors performs similarly in minFDE$_6$ and minADE$_6$ but significantly better in ASD$_6$ and FSD$_6$. 
This is because the behavior tokens guarantee that the generator will generate predictions for each driving direction equally. 
It is worth mentioning that the behavior tokens do not show a significant negative impact on the prediction accuracy measured in minFDE$_6$ and minADE$_6$ in comparison with the baseline model.
However, after removing the map token in CDT-B, though the diversity (ASD$_6$ and FSD$_6$) of the prediction slightly increases due to the lift of scene constraints, both the prediction accuracy (minFDE$_6$ and minADE$_6$) and scene compliance (ECFL$_6$) drop evidently, indicating that the map information is critical to achieve accurate and scene-compliant trajectory prediction.
Moreover, if we incorporate an accurately predicted endpoint, our model CDT-P can achieve excellent prediction accuracy measured in minFDE$_6$ and minADE$_6$ and also perfect scene compliance (ECFL$_6=0.99$).
Interestingly, in this case, the prediction diversity is very low. 
This is because the endpoint has largely contained the predicted trajectories within in a small range.
\begin{table}[t!]
\caption{Comparison of trajectory diversity and scene compliance in the validation set. Best values are in boldface.}
\begin{center}
\setlength{\tabcolsep}{3pt}
\begin{tabular}{l|ccccc}
\toprule
\textbf{Method}&\textbf{minFDE$_6$}&\textbf{minADE$_6$} & \textbf{ECFL$_6$} &\textbf{ASD$_6$}&\textbf{FSD$_6$}\\
\midrule
% QCNet\cite{zhou2023query} & 1.19 & 0.62 & - & 8.45 & 8.46 \\
CDT (Baseline) & 2.42 &  1.13 & 0.96 & 11.33 & 18.02\\
CDT w/o map token & 4.26 & 2.12 & 0.91 & \textbf{14.67}  & \textbf{42.30}\\
CDT-B & 2.45 & 1.13 & 0.97 & 14.49 & 40.75\\
CDT-P & \textbf{1.14} & \textbf{0.71} & \textbf{0.99} & 0.88 & 0.37\\
\bottomrule
\end{tabular}
\label{tab3}
\end{center}
\end{table}

\begin{figure*}[!t]
\centerline{\includegraphics[width=\linewidth, trim={0 0 0 0cm},clip]{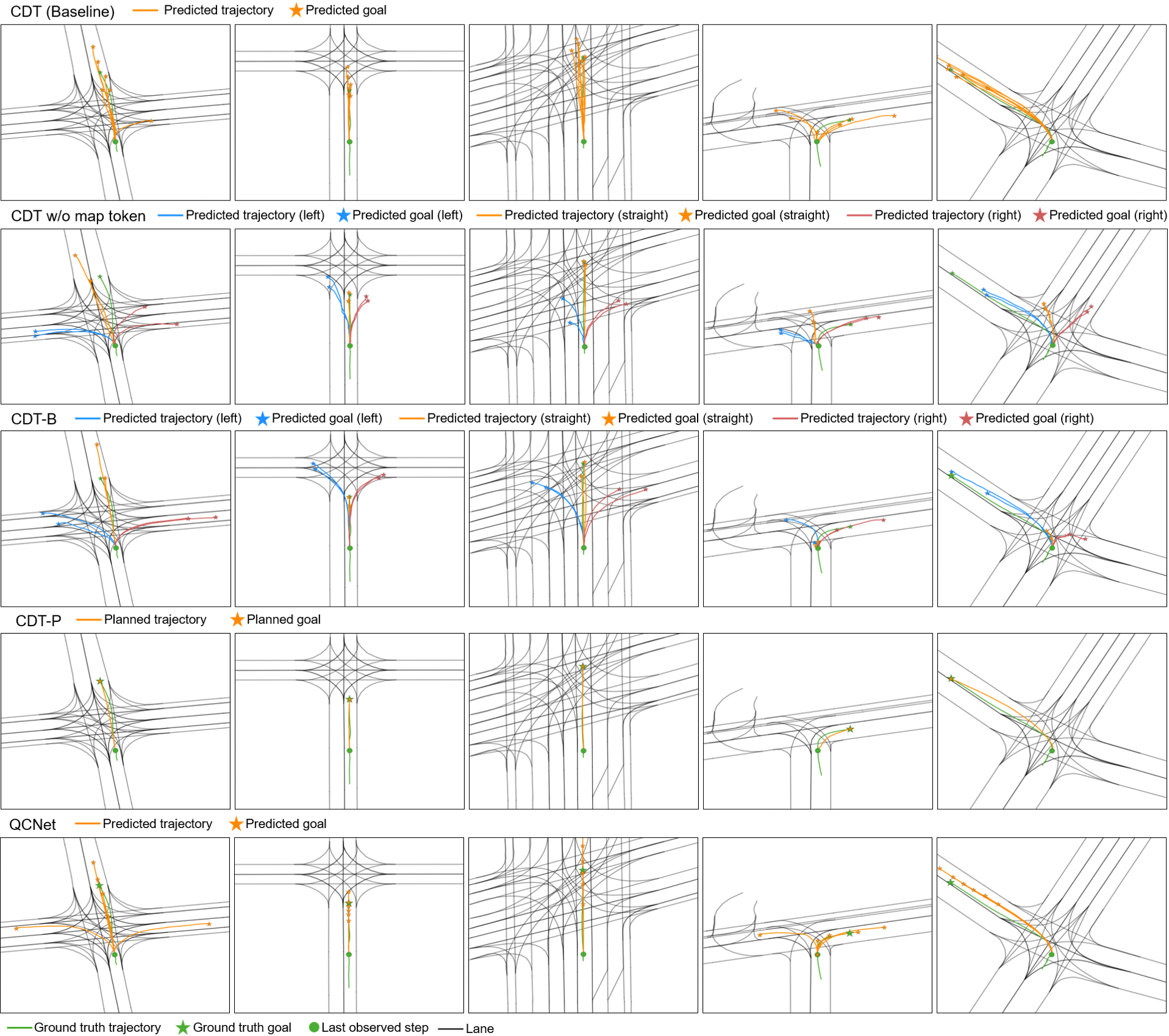}}
\caption{Qualitative comparison of the models under complex scenarios in the validation set.
Each column represents a unique intersection and each row represents the results predicted by the same model. 
The models include: 
CDT, the baseline model without any behavior tokens;
CDT w/o map token, the model has behavior tokens but without map token.
CDT-B, the model has both behavior and map token;
CDT-P, the model has endpoint and map token;
QCNet: the winner model on the Argoverse~2 leaderboard.}
\vspace{-6pt}
\label{qualitaty}
\end{figure*}

\vspace{3pt}
\noindent\textbf{Qualitative results.}
Furthermore, in Fig.~\ref{qualitaty}, we show the qualitative results of our proposed model for predicting trajectories in various driving scenes.
It can be clearly seen that the predictions of the baseline model CDT without behavior tokens are more homogeneous with respect to driving directions,
whereas CDT-B generates both more diverse and scene-consistent trajectories, suggesting two trajectories for each direction, i.e., turning left, going straight, and turning right. 
After removing the map information, despite the fact that CDT w/o map token maintains a good diversity, it sometimes even generates trajectories that are outside the driving areas or not aligned with the driving lanes.
After giving a predicted goal as the endpoint token to CDT-B, it generates very accurate trajectories that are well overlaid with the ground truth trajectories.
% As shown in Fig. \ref{qualitaty}, our trajectory prediction method, leveraging a conditional diffusion model with behavior control, adeptly generates multimodal sampling outcomes based on lane lines. Compared to other models, ours yields more diverse results. When provided with specified destination conditions, our model effectively plans intermediate trajectories along the lane lines and accurately reaches the designated endpoints. This capability underscores the model's proficiency in integrating lane line information for enhanced prediction diversity and planning accuracy in autonomous driving tasks.

\vspace{3pt}
\noindent\textbf{Comparison with state-of-the-art models.}
\begin{table}[!t]
\caption{Comparison with state-of-the-art models on the test set. Best and second best values are in boldface and underlined, respectively.}
\begin{center}
\setlength{\tabcolsep}{4.6pt}
\begin{tabular}{l|ccccc}
\toprule
\textbf{Method}&\textbf{minFDE$_6$}&\textbf{minADE$_6$}&\textbf{minMR$_6$}\\
\midrule
THOMAS\cite{gilles2021thomas} & 1.51 & 0.88 & 0.20 \\
MacFormer\cite{feng2023macformer} & 1.38 & 0.70 & 0.19  \\ 
ProphNet\cite{wang2023prophnet} & 1.32 & \underline{0.66} & 0.18  \\ 
Gnet\cite{gao2023dynamic}  & 1.34 & 0.69 & 0.18  \\
QCNet\cite{zhou2023query} & \textbf{1.19} & \textbf{0.62} &\textbf{ 0.14} & \\
\midrule
% CDT (Ours) & 2.95 & 1.33 & 0.41 & 11.33 & 18.02 \\
% CDT-B (Ours) & 2.97 &  1.33 & 0.47 & 14.49 &  40.75 \\
CDT-P (Ours) & \underline{1.25} &  0.80 & \underline{0.16}  \\
\bottomrule
\end{tabular}
\label{tab11}
\end{center}
\end{table}
In Table \ref{tab11}, we show the comparison of CDT-B with the state-of-the-art models in terms of prediction accuracy from the Argoverse~2 online test set. 
It can be seen that CDT-P is the runner-up model, outperforming most of the other models, such as THOMAS\cite{gilles2021thomas} and MacFormer\cite{feng2023macformer} measured in minFDE$_6$ and minMR$_6$.
Our approach slightly falls behind QCNet\cite{zhou2023query}.
The advantage of CDT-P lies in the generation of controllable predictions given by the endpoint token and mainly focuses on denoising the intermediate positions of the trajectory. 
Qualitative results on the validation are shown in Fig.~\ref{qualitaty}, in this case, CDT-P scarifies diversity to the alignment with the endpoint of ground truth, while QCNet is not constrained by the endpoint and generates more diverse predictions. 

\subsection{Further Analysis and Ablation Studies}
\begin{figure}[t!]
\centerline{\includegraphics[width=5.5cm]{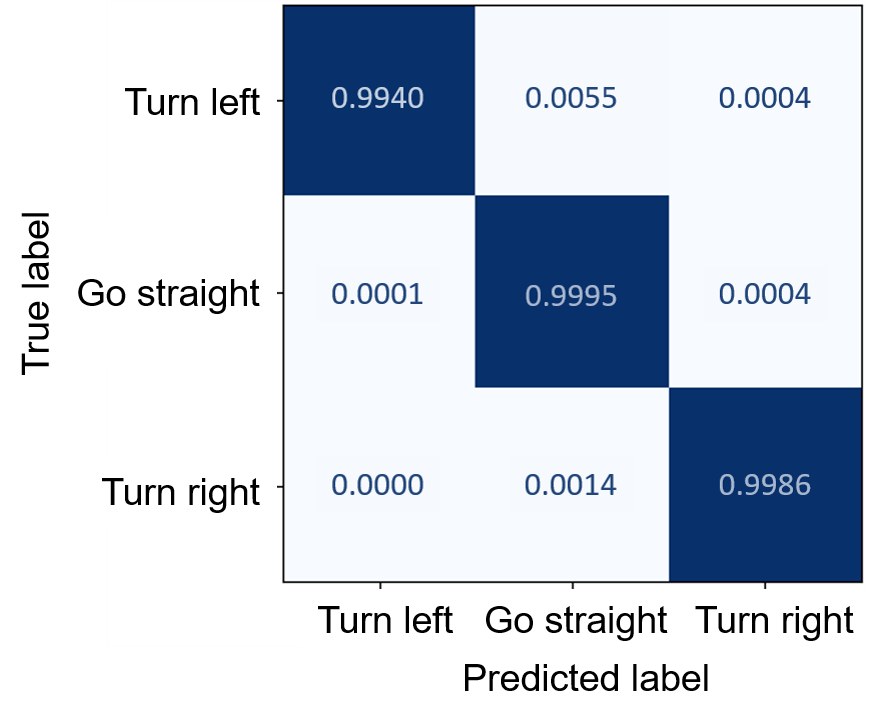}}
\caption{Normalized Confusion matrix of behavioral classification results in the validation set.}
\vspace{-6pt}
\label{mc}
\end{figure}

% \vspace{3pt}
\noindent\textbf{Analysis of the mode classifier.}
% In our study, we designed a classification head to categorize trajectory features from the encoder into behaviors such as going straight, turning left, and turning right. 
Due to the lack of ground truth data of the online test on Argoverse~2, we tested the performance of the mode classifier on a validation set.
It contains 20,469, 2,346, and 2,173 counts for straight, left turn, and right turn behaviors, respectively. 
We calculated the F1-Score, precision, recall, and accuracy for the classification performance.
The F1-Score, precision, recall, and accuracy found are 0.9977, 0.9979, 0.9974, and 0.9989. 
Fig.~\ref{mc} displays the confusion matrix for the behavior prediction results. 
Owing to the exceptional performance of the encoder, the classification results were outstanding, accurately predicting the probability of different behavioral modes. 
This precision in behavior classification is critical for enhancing the overall accuracy and reliability of the motion prediction framework.

\vspace{3pt}
\noindent\textbf{Ablation study of the diffusion model.}
We conducted ablation experiments also on the Argoverse~2 validation set. 
Our focus was on the number of denoising steps in the conditional diffusion model, which directly impacts the quality and efficiency of trajectory generation. Given that generating trajectory sequences (unlike two-dimensional images, which often require thousands denoising steps) is simpler and requires fewer denoising steps, we aim to validate our hypothesis through two sets of comparative experiments. In the first set, we trained networks with 5, 10, 20, 32, 50, and 100 diffusion steps for the same 100 epochs, comparing the quality of the generated trajectories. 
We found that the model starts to saturate with increased ADE and FDE when it is trained more than 50 epochs.
In the second set, we linearly increased the training epochs for different diffusion steps, setting it at seven times the number of diffusion steps (a value chosen to balance experimental fairness and training duration), resulting in 35, 70, 140, 224, and 350 epochs, respectively. 
% Due to the long training time, we did not test the results of 100 denoising steps after training for 700 epochs. 
This linear increase corresponds to the linear growth in the number of intermediate processes required with more diffusion steps. 

\begin{figure}[t!]
\centerline{\includegraphics[width=8.5cm]{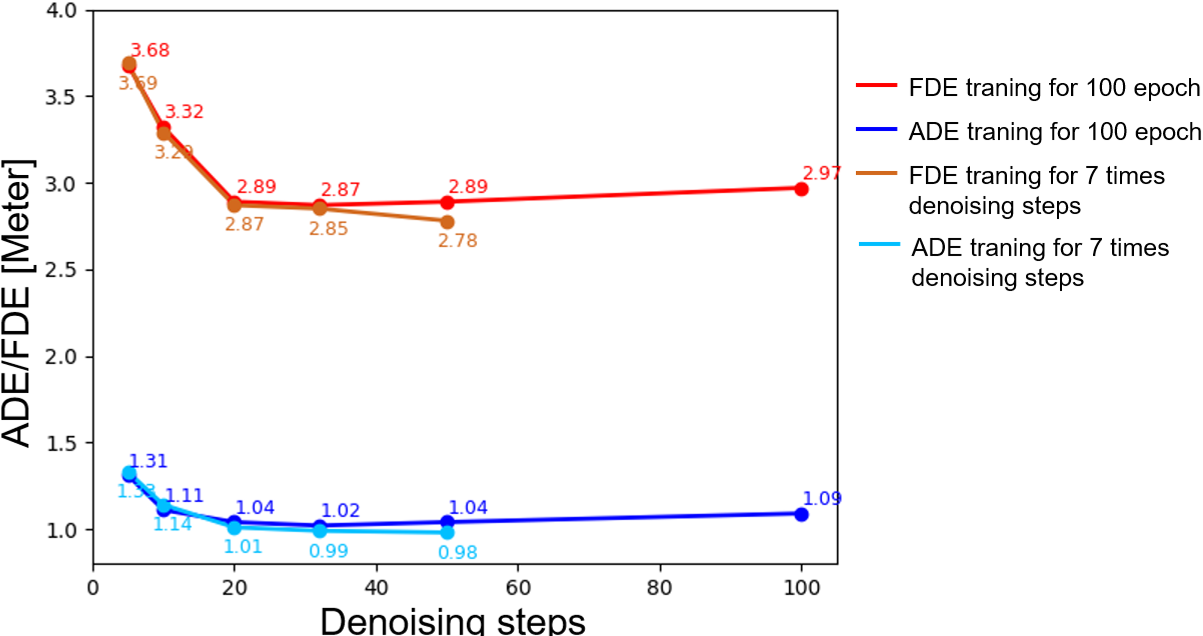}}
\caption{The network was trained with different numbers of denoising steps.}
\vspace{-6pt}
\label{DenoisingSteps}
\end{figure}

Based on the quantitative results presented in Fig.~\ref{DenoisingSteps},
we draw two conclusions: First, more diffusion steps necessitate longer training durations; Second, although increased diffusion steps improve performance, the marginal gains are limited and not cost-effective in terms of computational resources. Therefore, considering the trade-off between computational resource expenditure and effectiveness, we opted for 20 denoising steps and 140 epochs in our diffusion model.

% \begin{figure}[t!]
% \centerline{\includegraphics[width=9cm]{ablation.png}}
% \caption{The qualitative results are presented for three model variants: CDT-B, which is the complete model controlled by behavioral conditions; and CDT w/o behavior, which lacks the behavior control conditions; CDT w/o map, where the map token is removed.}
% \label{ablation}
% % \end{figure}
% In Table \ref{tab3}, we validated our approach to trajectory generation based on behavioral conditions. Removing the behavior control condition token not only resulted in an inability to steer the trajectory generation but also significantly lowered the prediction results. We also experimented by eliminating the map constraints, relying solely on the agents' historical trajectories (retaining the A-L module) for generation. A decline in the ECFL metric indicates a significant increase in the probability of generated trajectories colliding with non-drivable areas. This led to a significant decrease in prediction accuracy and resulted in the generation of trajectories that were inconsistent with actual map conditions, as shown in Fig. \ref{ablation}. This underscores the importance of both behavioral conditions and map constraints in accurate and realistic trajectory prediction, particularly in the context of trajectory prediction accuracy and adherence to real-world traffic scenarios.

\section{Conclusion and Discussion}
In this work, we introduce a novel motion prediction framework based on conditional diffusion models, capable of controlling motion behavior to generate multimodal trajectories that align with map environments. Through the empirical studies on the Argoverse~2 benchmark, we demonstrated the controllability of trajectory modalities based on behavioral conditions. Even in the case of data imbalance, controlling behavior enables the generative model to diversely sample different trajectory modalities. 
% We showcased advanced predictive results. 
In our ablation studies, we validated the effectiveness of the modules in our proposed method and the rationale behind our choice of denoising steps. 

\vspace{3pt}
\noindent\textbf{Limitation.} In this work, due to insufficient data samples for less common driving behaviors, such as U-turn, we simplified the driving behaviors to turning left, going straight, and turning right.
Also, the attributes of agents, such as cyclist type (motor, bicycle) and vehicle type (car, bus), are not considered in this work.
For future work, incorporating a wider range of control conditions with more comprehensive driving behaviors and sampling methods could further enhance the conditional diffusion model's trajectory quality and prediction efficiency. Furthermore, we want to include information about agent attributes in the agent token to guide trajectory prediction. This is because vehicles behave differently when driving past different agents, for example, a vehicle driver should be more cautious and slow down the driving speed when passing a group of playing children.

\end{document}